\documentclass[11pt,a4paper]{article}
\usepackage[hyperref]{acl2019}
\usepackage{times}
\usepackage{latexsym}
\usepackage{epsfig}
\usepackage{graphicx}
\usepackage{amsmath}
\usepackage{amssymb}
\usepackage{verbatim}
\usepackage{algorithm}
\usepackage[noend]{algpseudocode}
\usepackage{bm,color}
\usepackage{xcolor}
\usepackage{colortbl}
\definecolor{high}{HTML}{FFCE8E}

\usepackage{url}

\aclfinalcopy 

\title{What Should I Ask? Using Conversationally Informative Rewards for Goal-Oriented Visual Dialogue}

\author{Pushkar Shukla$^1$, Carlos Elmadjian$^{2}$, Richika Sharan$^ {1}$, Vivek Kulkarni$^{3}$,\\
\textbf{William Yang Wang$^{1}$, Matthew Turk$^{1}$} \\
  $^{1}$University of California, Santa Barbara\\
  $^{2}$University of S\~ao Paulo \\
  $^{3}$Stanford University\\
  $^{1}$\texttt{\{pushkarshukla,richikasharan,wangwilliamyang,mturk\}@ucsb.edu} \\
 $^{2}$ \texttt{elmad@ime.usp.br}\\
 $^{3}$ \texttt{viveksck@stanford.edu}
  }

\date{}

\begin{document}
\maketitle
\begin{abstract}
 The ability to engage in goal-oriented conversations has allowed humans to gain knowledge, reduce uncertainty, and perform tasks more efficiently. Artificial agents, however, are still far behind humans in having goal-driven conversations. In this work, we focus on the task of goal-oriented visual dialogue, aiming to automatically generate a series of questions about an image with a single objective. This task is challenging, since these questions must not only be consistent with a strategy to achieve a goal, but also consider the contextual information in the image. We propose an end-to-end goal-oriented visual dialogue system,  that combines reinforcement learning with regularized information gain. Unlike previous approaches that have been proposed for the task, our work is motivated by the Rational Speech Act framework, which models the process of human inquiry to reach a goal. We test the two versions of our model on the GuessWhat?!~dataset, obtaining significant results that outperform the current state-of-the-art models in the task of generating questions to find an undisclosed object in an image. 
\end{abstract}

\section{Introduction}
\begin{figure}[ht]
	\centering
	\includegraphics[,width=0.94\columnwidth]{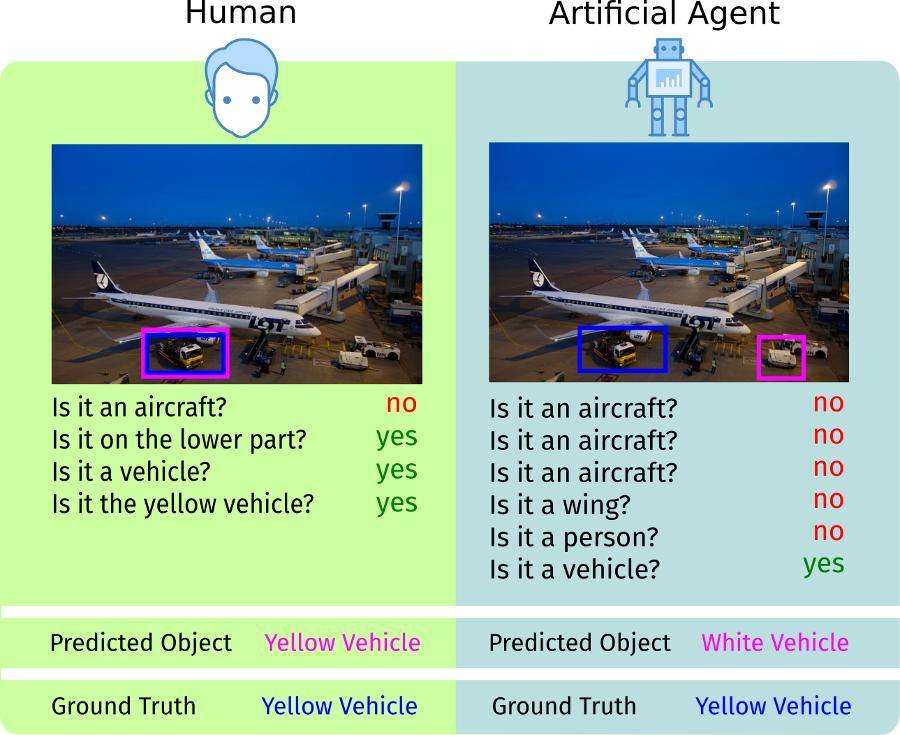}
	\caption{
	An example of goal-oriented visual dialogue for finding an undisclosed object in an image through a series of questions. On the left, we ask a human to guess the unknown object in the image. On the right, we use the baseline model proposed by Strub et al.~\cite{strub2017end}. While the human is able to narrow down the search space relatively faster, the artificial agent is not able to adopt a clear strategy for guessing the object.}~\label{human_vs_ai}
\end{figure}

Building natural language models that are able to converse towards a specific goal is an active area of research that has attracted a lot of attention in recent years. These models are vital for efficient human-machine collaboration, such as when interacting with personal assistants. In this paper, we focus on the task of goal-oriented visual dialogue, which requires an agent to engage in conversations about an image with a predefined objective. The task presents some unique challenges. Firstly, the conversations should be consistent with the goals of the agent. Secondly, the conversations between two agents must be coherent with the common visual feedback. Finally, the agents should come up with a strategy to achieve the objective in the shortest possible way. This is different from a normal dialogue system where there is no constraint on the length of a conversation.

Inspired by the success of Deep Reinforcement Learning, many recent works have also used it for building models for goal-oriented visual dialogue \cite{bordes2016learning}. The choice makes sense, as reinforcement learning is well suited for tasks that require a set of actions to reach a goal. However, the performance of these models have been sub-optimal when compared to the average human performance on the same task. For example, consider the two conversations shown in Figure~\ref{human_vs_ai}. The figure draws a comparison between possible questions asked by humans and an autonomous agent proposed by Strub et al.~\cite{strub2017end} to locate an undisclosed object in the image. While humans tend to adopt strategies to narrow down the search space, bringing them closer to the goal, it is not clear whether an artificial agent is capable of learning a similar behavior only by looking at a set of examples. This leads us to pose two questions: \textit{What strategies do humans adopt while coming up with a series of questions with respect to a goal?}; and \textit{Can these strategies be used to build models that are suited for goal-oriented visual dialogue?}

With this challenge in mind, we directed our attention to contemporary works in the field of cognitive science, linguistics and psychology for modelling human inquiry~\cite{groenendijk1984semantics, nelson2005finding,van2003questioning}. More specifically, our focus lies on how humans come up with a series of questions in order to reach a particular goal. One popular theory suggests that humans try to maximize the expected regularized information gain while asking questions~\cite{hawkins2015you,coenen2017asking}. Motivated by that, we evaluate the utility of using information gain for goal-oriented visual question generation with a reinforcement learning paradigm. In this paper, we propose two different approaches for training an end-to-end architecture: first, a novel reward function that is a trade-off between the expected information gain of a question and the cost of asking it; and second, a loss function that uses regularized information gain with a step-based reward function. Our architecture is able to generate goal-oriented questions without using any prior templates. Our experiments are performed on the GuessWhat?! dataset~\cite{de2017guesswhat}, a standard dataset for goal-oriented visual dialogue that focuses on identifying an undisclosed object in the image through a series of questions. Thus, our contribution is threefold:

\begin{itemize}
    \item An end-to-end architecture for goal-oriented visual dialogue combining Information Gain with Reinforcement Learning.
    \item A novel reward function for goal-oriented visual question generation to model long-term dependencies in dialogue.
    \item Both versions of our model outperform the current baselines on the GuessWhat?! dataset for the task of identifying an undisclosed object in an image by asking a series of questions. 
    
\end{itemize}



\section{Related Work}

%
\subsection{Models for Human Inquiry}

There have been several works in the area of cognitive science that focus on models for question generation. Groenendijk et al.~\cite{groenendijk1984semantics} proposed a theory stating that meaningful questions are propositions conditioned by the quality of its answers. Van Rooy~\cite{van2003questioning} suggested that the value of a question is proportional to the questioner's interest and the answer that is likely to be provided. Many recent related models take into consideration the optimal experimental design (OED)~\cite{nelson2005finding,gureckis2012self}, which considers that humans perform intuitive experiments to gain information, while others resort to Bayesian inference. Coenen et al.~\cite{coenen2017asking}, for instance, came up with nine important questions about human inquiry, while one recent model called Rational Speech Act (RSA) \cite{hawkins2015you} considers questions as a distribution that is proportional to the trade-off between the expected information gain and the cost of asking a question. 

\begin{figure*}[h]
	\centering
	\includegraphics[width=0.8\textwidth]{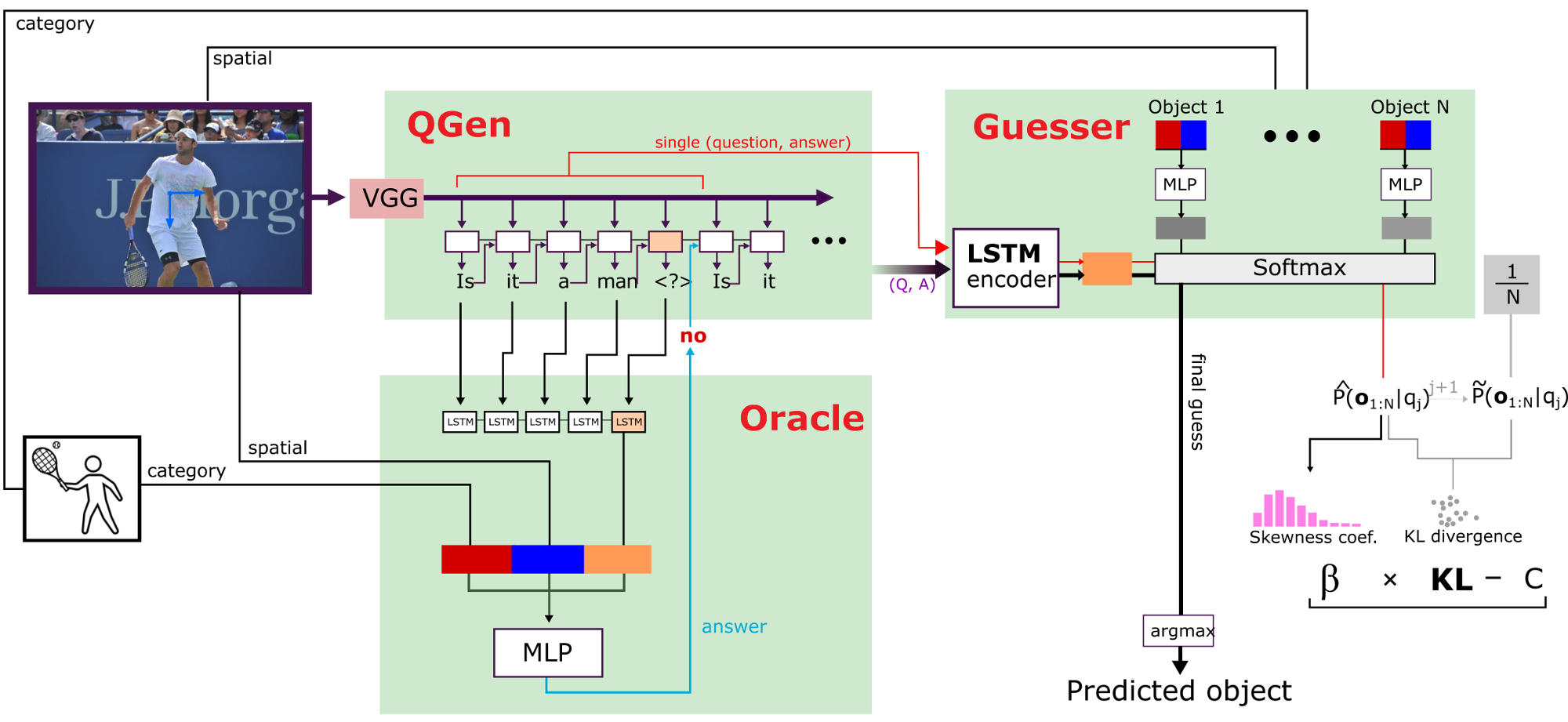}
	\caption{A block diagram of our model.
	 The framework is trained on top of three individual models: the questioner (QGen), the guesser, and the oracle. The guesser returns an object distribution given a history of question-answer pairs that are generated by the questioner and the oracle respectively. These distributions are used for calculating the information gain of the question-answer pair. The information gain and distribution of probabilities given by the Guesser 
	 are
	used either as a reward or optimized as a loss function with global rewards for training the questioner.}~\label{framework}
\end{figure*} 
\subsection{Dialogue Generation and Visual Dialogue}
Dialogue generation is an important research topic in NLP, thus many approaches have been proposed to address this task.  Most earlier works made use of a predefined template~\cite{lemon2006isu,wang2013simple} to generate dialogues. More recently, deep neural networks have been used for building end-to-end architectures capable of generating questions~\cite{vinyals2015neural,sordoni2015neural} and also for the task of goal-oriented dialogue generation~\cite{rajendran2018learning,bordes2016learning}.
Visual dialogue  focuses on having a conversation about an image with either one or both of the agents being a machine. Since its inception~\cite{das2017visual}, different approaches have been proposed to address this problem~\cite{massiceti2018flipdial,lu2017best,das2017visual}.
Goal-oriented Visual Dialogue, on the other hand, is an area that has only been introduced fairly recently. De Vries et al.~\cite{de2017guesswhat} proposed the GuessWhat?! dataset for goal-oriented visual dialogue while Strub et al.~\cite{strub2017end} developed a reinforcement learning approach for goal-oriented visual question generation. More recently, Zhang et al.~\cite{zhang2018goal} used intermediate rewards for training a model on this task.

\subsection{Sampling Questions with Information Gain}
  Information gain has been used before to build question-asking agents, but most of these models resort to it to sample questions. Rothe et al.~\cite{rothe2017question} proposed a model that generates questions in a Battleship game scenario. Their model uses Expected Information Gain to come up with questions akin to what humans would ask. Lee et al.~\cite{lee2018answerer} used information gain alone to sample goal-oriented questions on the GuessWhat?! task in a non-generative fashion. The most similar work to ours was proposed by Lipton et al.~\cite{lipton2017bbq}, who used information gain and Q-learning to generate goal-oriented questions for movie recommendations. However, they generated questions using a template-based question generator.

\section{The GuessWhat?! framework}
We built our model based on the GuessWhat?! framework \cite{de2017guesswhat}. GuessWhat?! is a two-player game in which both players are given access to an image containing multiple objects. One of the players -- the \textit{oracle} -- chooses an object in the image. The goal of the other player -- the \textit{questioner} -- is to identify this object by asking a series of questions to the oracle, who can only give three possible answers: "yes," "no," or "not applicable." Once enough evidence is collected, the questioner has to choose the correct object from a set of possibilities -- which, in the case of an artificial agent, are evaluated by a \textit{guesser} module. If this final guess is correct, the questioner is declared the winner.  The GuessWhat?! dataset comprises 155,280 games on 66,537 images from the MS-COCO dataset, with 831,889 question-answer pairs. The dataset has 134,074 unique objects and 4,900 words in the vocabulary. 

A game is comprised of an image $I$ with height $H$ and width $W$, a dialogue $D = \{({q_1},{a_1}),({q_2},{a_3}),...({q_n},{a_n})$\}, where ${q_j}$ $\in$ ${Q}$ denotes a question from a list of questions and ${a_j}$ $\in$ ${A}$ denotes an answer from a list of answers, which can either be $\langle$yes$\rangle$, $\langle$no$\rangle$ or $\langle$N/A$\rangle$. The total number of objects in the image is denoted by $O$ and the target is denoted by $o^*$. 
The term $V$ indicates the vocabulary that comprises all the words that are employed to train the question generation module (QGen). Each question can be represented by $q=\{w^i\}$, where $w^i$ denotes the $i^{th}$ word in the vocabulary. The set of segmentation masks of objects is denoted by $S$. These notations are similar to those of Strub et al.~\cite{strub2017end}. An example of a game can be seen in Figure 1, where the questioner generates a series of questions to guess the undisclosed object. In the end, the guesser tries to predict the object with the image and the given set of question-answer pairs. 

\subsection{Learning Environment}
We now describe the preliminary models for the questioner, the guesser, and the oracle. 
Before using them for the GuessWhat?! task, we pre-train all three models in a supervised  manner. During the final training of the Guesswhat?! task our focus is on building a new model for the questioner and we use the existing pre-trained models for the oracle and the guesser.

\subsubsection{The Questioner}

The questioner's job is to generate a new question $q_{j+1}$ given the previous $j$ question-answer pairs and the image $I$. Our model has a similar architecture to the VQG model proposed by Strub et al.~\cite{strub2017end}. It consists of an LSTM whose inputs are the representations of the corresponding image $I$ and the input sequence corresponds to the previous dialogue history. The representations of the image are extracted from the \textit{fc-8} layer of the VGG16 network~\cite{simonyan2014very}. The output of the LSTM is a probability distribution over all words in the vocabulary. The questioner is trained in a supervised fashion by minimizing the following negative log-likelihood loss function: 
\begin{multline}
L_{ques} = -\log_{p_q}(q_{1:J}|I,a_{1:J})\\
=-\sum_{j=1}^J\sum_{i=1}^{I_{j}}\log_{p_q}(w_i^j|w_{1:i-1}^j,(q,a)_{1:j-1},I)
\end{multline}

Samples are generated in the following manner during testing: given an initial state $s_{0}$ and new token $w_{0}^{j}$, a word is sampled from the vocabulary. The sampled word along with the previous state is given as the input to the next state of the LSTM. The process is repeated until the output of the LSTM is the $\langle$end$\rangle$ token.

\subsubsection{The Oracle}
The job of the oracle is to come up with an answer  to each question that is posed. In our case, the three possible outcomes are $\langle$yes$\rangle$, $\langle$no$\rangle$, or $\langle$N/A$\rangle$. The architecture of the oracle model is similar to the one proposed by De Vries et al.~\cite{de2017guesswhat}. The input to the oracle is an image, a category vector, and the question that is encoded using an LSTM. The model then returns a distribution over the possible set of answers.

\subsubsection{The Guesser}
The job of the guesser is to return a distribution of probabilities over all set of objects given the input image and the dialogue history. We convert the entire dialogue history into a single encoded vector using an LSTM. All objects are embedded into vectors, and the dot product of these embeddings are performed with the encoded vector containing the dialogue history. The dot product is then passed through an MLP layer that returns the distribution over all objects.

\section {Regularized Information Gain}
The motivation behind using Regularized Information Gain (RIG) for goal-oriented question-asking comes from the Rational Speech Act Model (RSA)~\cite{hawkins2015you}. RSA tries to mathematically model the process of human questioning and answering. According to this model, when selecting a question from a set of questions, the questioner considers a goal $g$ $\in$ $G$ with respect to the world state $G$ and returns a probability distribution of questions such that:

\begin{equation} \label{eq:prob-questions}
    P(q|g)\propto e^{D_{KL}(\overset{\wedge}{p}(q|g)||\overset{\sim}{p}(q|g)) - C(q)}\vspace{0.2cm}
\end{equation}
where $P(q|g)$ represents probability of selecting a question $q$ from a set of questions $Q$. The probability is directly proportional to the trade-off between the cost of asking a question $C(q)$ and the expected information gain $D_{KL}(\overset{\sim}{p}(q|g)||\overset{\wedge}{p}(q|g))$. The cost may depend on several factors such as the length of the question, the similarity with previously asked questions, or the number of questions that may have been asked before. The information gain is defined as the KL divergence between the prior distribution of the world with respect to the goal, $\overset{\sim}{p}(q|g)$, and the posterior distribution that the questioner would expect after asking a question, $\overset{\wedge}{p}(q|g)$. 

Similar to Equation \ref{eq:prob-questions}, in our model we make use of the trade-off between expected information gain and the cost of asking a question for goal-oriented question generation. Since the cost term regularizes the expected information gain, we denote this trade-off as Regularized Information Gain. For a given question \textit{q}, the Regularized Information Gain is given as:
\begin{equation}
\small
RIG (q) =
\tau(q) -C(q))
\end{equation}
where $\tau(q)$ is the expected information gain associated with asking a question and $C(q)$ is the cost of asking a question $q \in Q$ in a given game. Thus, the information gain is measured as the KL divergence between the prior and posterior likelihoods of the scene objects before and after a certain question is made, weighted by a skewness coefficient $\beta(q)$ over the same posterior. 
\begin{equation}
\small 
    \tau(q)= D_{KL}(\overset{\sim}{p}(q_{j}|I,(q,a)_{1:j-1})||\overset{\wedge}{p}(q_{j}|I,(q,a)_{1:j-1}))\beta(q)
\end{equation}

The prior distribution before the start of the game is assumed to be $\frac{1}{N}$, where N is the total number of objects in the game. After a question is asked, the prior distribution is updated and it is equal to the output distribution of the guesser:
\begin{equation}
    \footnotesize
    \overset{\sim}{p}(q_{j}|I,(q,a)_{1:j-1}=
        \begin{cases}
            p_{guess}(I,(q,a)_{1:j-1}),& \text{if $i \ge 1$}\\
            \frac{1}{N} ,& \text{if $i = 0$}
        \end{cases}
\end{equation}

We define the posterior to be the output of the guesser once the answer has been given by the oracle: 
\begin{equation}
    \overset{\wedge}{p}(q_j|I,(q,a))_{1:j-1})=\sum_{a\in A}^{A}p_{guess}(q_{j}|I,(q,a)_{1:j-1})
\end{equation}

The idea behind using skewness is to reward questions that lead to a more skewed distribution at each round. The implication is that a smaller group of objects with higher probabilities lowers the chances of making a wrong guess by the end of the game. Additionally, the measure of skewness also works as a counterweight to certain scenarios where KL divergence itself should not reward the outcome of a question, such as when there is a significant information gain from a previous state but the distribution of likely objects, according to the guesser, becomes mostly homogeneous after the question.

Since we assume that initially all objects are equally likely to be the target, the skewness approach is only applied after the first question. We use the posterior distribution provided by the guesser to extract the Pearson's second skewness coefficient (i.e., the \textit{median skewness}) and create the $\beta$ component. Therefore, assuming a sample mean $\mu$, median $m$, and standard deviation $\sigma$, the skewness coefficient is simply given by:
\begin{equation}
    \beta(q) =  \frac{3(\mu - m)}{\sigma}
\end{equation}

Some questions might have a high information gain, but at a considerable cost. The term $C(q)$ acts as a regularizing component to information gain and controls what sort of questions should be asked by the questioner. The cost of asking a question can be defined in many ways and may differ from one scenario to another. In our case, we are only considering whether a question is being asked more than once, since a repeated question cannot provide any new evidence that will help get closer to the target, despite a high information gain from one state to another during a complete dialogue. The cost for a repeated question is defined as:
\begin{equation}
C(q)= 
    \begin {cases}
        \tau(q),&  \text{if $q_j \in \{q_{j-1},...,q_1\}$}\\
        0, & \text{otherwise}
    \end{cases}
\end{equation}

The cost for a question is equal to the negative information gain. This sets the value of an intermediate reward to 0 for a repeated question,  ensuring that the net RIG is zero when the question is repeated.

\section {Our Model}
We view the task of generating goal-oriented visual questions as a Markov Decision Process (MDP), and we optimize it using the Policy Gradient algorithm. In this section, we  describe some of the basic terminology employed in our model before moving into the specific aspects of it.

  At any time instance $t$, the state of the agent can be written as $\bm{u_t} = ((w_1^j,...,w_m^j),(q,a)_{1:j-1},I)$, where $I$ is the image of interest, $(q,a)_{1:j-1}$ is the question-answer history, and $(w_1^j,...,w_m^j)$ is the previously generated sequence of words for the current question $q_j$. The action $v_t$ denotes the selection of the next output token from all the tokens in the vocabulary. All actions can lead to one of the following outcomes:
\begin{enumerate}
    \item The selected token is $\langle stop \rangle$, marking the end of the dialogue. This shows that it is now the turn of the guesser to make the guess.
    \item The selected token is $\langle end \rangle$, marking the end of a question. 
    \item The selected token is another word from the vocabulary. The word is then appended to the current sequence $(w_1^j,...,w_m^j)$. This marks the start of the next state.
\end{enumerate}
\begin{algorithm}
\caption{Training the question generator using REINFORCE with the proposed rewards}\label{alg:KL}
\begin{algorithmic}[1]
\small
\Require{Pretrained QGen, Oracle and Guesser}
\Require{Batch size $K$}
    \For{Each update}
        \For{$k = 1$ to $K$}
            \State{Pick image $I_k$ and the target object $o_k^* \in O_k$}
            \State{$N \gets |O_k|$}
            \State{$\overset{\sim}p(o_{k_{1:N}}) \gets \frac{1}{N}$}
            \For{$j \gets 1$ to $J_{max}$}
                \State{$q_j^k \gets QGen((q,a)_{1:j-1}^k, I_k$)}
                \State{$a_j^k \gets Oracle(q_j^k, o_k^*, I_k)$}
                \State{$\overset{\wedge}p(o_{k_{1:N}}) \gets Guesser((q,a)_{1:j}^k, I_k, O_k)$}
                \State{$\beta(q_j^k) \gets Skewness(\overset{\wedge}{p}(o_{k_{1:N}}))$}
                \State{$\tau(q_j^k) \gets D_{KL}(\overset{\sim}{p}(o_{k_{1:N}})||\overset{\wedge}{p}(o_{k_{1:N}})) \beta(q_j^k)$}
                \State{$C(q_j^k) \gets \begin{cases}
                        \tau(q_j^k) & \text{if } q_j \in \{q_{j-1},...,q_1\}\\
                        0 & \text{Otherwise}
                    \end{cases}$}
               
            \EndFor
            \State{$R \gets  \sum_{j=1}^{J_{max}}\tau(q_j^k) - C(q_j^k)$}
            \State{$p(o_k|\cdot) \gets Guesser((q,a)_{1:j}^k, I_k, O_k)$}
            \State{$r(\bm{u_t},v_t) \gets \begin{cases}
                   R & \text{If argmax}\  p(o_k|\cdot)  = o_k^*\\
                   0 & \text{Otherwise}
            \end{cases}$}
        \EndFor
        \State{Define $T_h \gets ((q,a)_{1:j_k}^k, I_k, r_k)_{1:K}$}
        \State{Evaluate $\nabla J(\theta_h)$ with Eq.13 with $T_h$}
        \State{SGD update of QGen parameters $\theta$ using $\nabla J(\theta_h)$}
        \State{Evaluate $\nabla L(\phi_h)$ with Eq.15 with $T_h$}
        \State{SGD update of baseline parameters using $\nabla L(\phi_h)$}
    \EndFor
\end{algorithmic}
\label{algorithm}
\end{algorithm}

Our approach models the task of goal-oriented questioning as an optimal stochastic policy $\pi_{\theta}(v|\bm{u})$ over the possible set of state-action pairs. Here $\theta$ represents the parameters present in our architecture for question generation. In this work, we experiment with two different settings to train our model with Regularized Information Gain and policy gradients. In the first setting, we use Regularized Information Gain as an additional term in the loss function of the questioner. We then train it using policy gradients with a 0-1 reward function. In the second setting, we use Regularized Information Gain to reward our model. Both methods are described below.

\subsection{Regularized Information Gain loss minimization with 0-1 rewards}
During the training of the GuessWhat?! game we introduce Regularized Information Gain as an additional term in the loss function. The goal is to minimize the negative log-likelihood and maximize the Regularized Information Gain. The loss function for the questioner is given by: 
\begin{equation}
    \begin{split}
        \footnotesize
 L(\theta) = -\log_{p_q}(q_{1:J}|I,a_{1:J})+ \tau(q) -C(q) \\
 =-\sum_{j=1}^J\sum_{i=1}^{I_{j}}\log_{p_q}(w_i^j|w_{1:i-1}^j,(q,a)_{1:j-1},I)\\
 + D_{KL}(\overset{\wedge}{p}(q_{j}|I,(q,a))||\overset{\sim}{p}(q_{j}|I,(q,a)))\beta(q)
    \end{split}
\end{equation}

\begin{table*}
\centering
\small
        
            \begin{tabular}{|l|cccc||cccc|}
                \hline
                \textbf{} &\multicolumn{4}{ c ||}{\textbf{New Image}}&\multicolumn{4}{ c |}{\textbf{New Object}}\\
                \textbf{Approach}  &{\textbf{ Greedy }} &{\textbf{ Beam }} &{\textbf{ Sampling}}& {\textbf{ Best}} &{\textbf{ Greedy }} &{\textbf{ Beam }} & {\textbf{ Sampling }}& {\textbf{ Best}}\\
                
                \hline

                Baseline.\cite{strub2017end} 	& {46.9\%} & {53.0\%} & {45.0\%}&  {53.0\%}     & {53.4\%} & {46.4\%} & {46.44\%}&{53.4\%} \\

                Strub et al. \cite{strub2017end} &  {58.6\%} & {54.3\%} & {63.2\%}&{63.2\%}     &  {57.5\%} & {53.2\%} & {62.0\%}&{62.0\%}\\
                Zhang et al. ~\cite{zhang2018goal} & {56.1\%} &{54.9\%} &{55.6\%} &{55.6\%}& {56.51\%} &{56.53\%} &{49.2\%}&{56.53\%}\\
                $TPG^{1}$\cite{zhao2018learning}  &$-$&{}$-$&$-$&{62.6\%}&$-$&$-$&$-$&$-$\\

                GDSE-C \cite{venkatesh2018jointly}  &$-$&$-$&$-$&60.7\%&$-$&$-$&$-$&63.3\%\\
                $ISM^{1}$ \cite{abbasnejad2018active}  &$-$&$-$&$-$&62.1\%&$-$&$-$&$-$&64.2\%\\

                \rowcolor{high}
                RIG as rewards & {59.0\%} &{60.21\%} &{{64.06}\%}&{\textbf{64.06}\%}& {{63.00}\%} & {{63.08}\%}& {{65.20}\%}&{\textbf{65.20}\%}\\
                \rowcolor{high}
                {RIG as a loss  with 0-1 rewards}& {61.18\%} &{59.79\%} &{{65.79\%}}&\textbf{{65.79\%}}& {63.19\%} & {62.57\%} & { {67.19\%}}&{\textbf {67.19\%}}\\
                
                \hline
            
            \end{tabular}
    \caption{ A comparison of the recognition accuracy of our model with the state of the art model \cite{strub2017end} and other concurrent models  on the GuessWhat?! task for guessing an object in the images from the test set.} 
    \label{table1}
\end{table*}

We adopt a reinforcement learning paradigm on top of the proposed loss function. We use a zero-one reward function similar to Strub et al.~\cite{strub2017end} for training our model. The reward function is given as: 
\begin{equation}
\small
r(\bm{u_t},v_t)= 
    \begin {cases}
        \small 1 ,&  \text{if argmax}_{p_{guess}}=o^*\\
        0, & \text{otherwise}
    \end{cases}
\end{equation}
Thus, we give a reward of $1$ if the guesser is able to guess the right object and $0$ otherwise.

\subsection{Using Regularized Information Gain as a reward}
Defining a valuable reward function is a crucial aspect for any Reinforcement Learning problem. There are several factors that should be considered while designing a good reward function for asking goal-oriented questions. First, the reward function should help the questioner achieve its goal. Second, the reward function should optimize the search space, allowing the questioner to come up with relevant questions. 
The idea behind using regularized information gain as a reward function is to take into account the long term dependencies in dialogue. 
Regularized information gain as a reward function can help the questioner to come up with an efficient strategy to narrow down a large search space. The reward function is given by: 
\begin{equation}
\footnotesize
    r(\bm{u_t},v_t)= 
    \begin {cases}
        \sum_{j=1}^{|Q|}(\tau(q_j) -C(q_j)) ,&  \scriptsize{\text{if argmax}\ {p_{guess}}=o^*}\\
        0, & \scriptsize{\text{otherwise}}
    \end{cases}
\end{equation}
Thus, the reward function is the sum of the trade-off between the information gain $\tau(q)$ and the cost of asking a question $C(q)$ for all questions $Q$ in a given game. Our function only rewards the agent if it is able to correctly predict the oracle's initial choice.

\subsection{Policy Gradients}
Once the reward function is defined, we train our model using the policy gradient algorithm. For a given policy \textit{$\pi_{\theta}$}, the objective function of the policy gradient is given by: 
\begin{equation}
\small
J(\theta)= E_{\pi_{\theta}}\Bigg[\sum_{t=1}^T r(u_t,v_t)\Bigg]
\end{equation}
According to Sutton et al. \cite{sutton2000policy}, the gradient of $J(\theta)$ can be written as: 
\begin{equation}
\small
\nabla J(\theta) \approx  \Bigg \langle    \sum_{t=1}^{T} \sum_{v_{t} \in V}\nabla_{\theta}log_{\pi_{\theta}}(u_{t},v_{t})(Q^{\pi_{\theta}}(u_{t},v_{t})- b_{\phi}) \Bigg \rangle
\end{equation}
where $Q^{\pi_{\theta}}(u_{t},v_{t})$ is the state value function given by the sum of the expected cumulative rewards:

\begin{equation}
   \small Q^{\pi_{\theta}}(u_t,v_t)= E_{\pi_{\theta}}\Bigg[\sum_{t'=t}^{T} r(u_{t},v_{t})\Bigg]
\end{equation}

Here $b_{\phi}$ is the baseline function used for reducing the variance. The baseline function is a single-layered MLP that is trained by minimizing the squared loss error function given by:

\begin{equation}
\footnotesize
\text{min}\ L_{\phi}=\Big \langle[ b_{\phi} -r(u_{t},v_{t})]^{2}\Big \rangle
\end{equation}


\begin {comment}
\begin{table}

\begin{center}

\begin{tabular}{|l|c|}
\hline
\raisebox{0ex}{\textbf{Approach}} &\raisebox{0ex}{\textbf{ Accuracy } } \\
\hline
 
Baseline (Greedy) &	 {40.7\%}\ \\ 
Baseline (Beam Search)& {46.1\%}\\
Baseline (Sampling)	& {39.2\%} \\\hline
Strub et al.~\cite{strub2017end}(Greedy) &  {49.5\%} \\ 
Strub et al.~\cite{strub2017end}(Beam Search) & {44.9\%}\\ 
Strub et al.~\cite{strub2017end}(Sampling)&  {53.3\%}\\\hline
RIG as rewards(Greedy)& {53.18\%} \\
RIG as rewards l(Beam Search)& {52.92\%}\\
\textbf{RIG as rewards (Sampling)}& {\textbf{55.03}\%}\\\hline
RIG loss with 0-1 rewards(Greedy)& {53.34\%} \\
RIG loss with 0-1 rewards l(Beam Search)& {52.81\%}\\
\textbf{RIG loss with 0-1 rewards (Sampling)}& {\textbf 56.71\%}\\\hline

\hline

\end{tabular}
\end{center}
\caption{Statistical comparison of our model with other approaches for guessing uniformly sampled objects in the training set.}
\label{table2}
\end{table}
\end{comment}

\section{Results}
\begin {comment}
\begin{table}

\begin{center}
\scalebox{1}{
\begin{tabular}{|l|c|}
\hline
\raisebox{0ex}{\textbf{Approach}} &\raisebox{0ex}{\textbf{ Accuracy} } \\
\hline
 
Baseline (Greedy) &	 {39.4\%}\ \\ 
Baseline (Beam Search)& {44.8\%}\\
Baseline (Sampling)	& {38.0\%} \\\hline

Strub et al.~\cite{strub2017end}  (Greedy) &  {48.5\%} \\ 
Strub et al.~\cite{strub2017end}  (Beam Search) & {45.8\%}\\ 
Strub et al.~\cite{strub2017end}  (Sampling)&  {52.3\%} \\ \hline

Proposed Model (Greedy)& {49.72\%} \\
Proposed Model (BeamSearch)& {49.20\%} \\
\textbf{Proposed Model (Sampling)}&{\textbf{54.07\%}} \\ \hline
RIG loss with 0-1 rewards(Greedy)& {50.80\%} \\
RIG loss with 0-1 rewards l(Beam Search)& {50.47\%}\\
\textbf{RIG loss with 0-1 rewards (Sampling)}& {\textbf{55.53}\%}\\\hline

\end{tabular}}
\end{center}
\caption{ Statistical comparison of our model with other models for guessing an object in the images present in the test set.}
\label{table1}

\end{table}
\end{comment}

The model was trained under the same settings of~\cite{strub2017end}. This was done in order to obtain a more reliable comparison with the preexisting models in terms of accuracy. After a supervised training of the question generator, we ran our reinforcement procedure using the policy gradient for 100 epochs on a batch size of 64 with a learning rate of 0.001. The maximum number of questions was 8. The baseline model, the oracle, and the guesser were also trained with the same settings described by~\cite{de2017guesswhat}, in order to compare the performance of the two reward functions. The error obtained by the guesser and the oracle were 35.8\% and 21.1\%, respectively.
\footnote{In order to have a fair comparison, the results reported for TPG \cite{zhao2018learning} and \cite{abbasnejad2018active} only take into consideration the performance of the question generator. We do not report the scores that were generated after employing memory network to the guesser. }

Table~\ref{table1} shows our primary results along with the baseline model trained on the standard cross-entropy loss for the task of guessing a new object in the test dataset. We compare our model with the one presented by~\cite{strub2017end} and other concurrent approaches. Table~\ref{table1} also compares our model with others when objects are sampled using a uniform distribution (right column). 
\begin{figure*}
	\centering
	\includegraphics[width=0.95\textwidth]{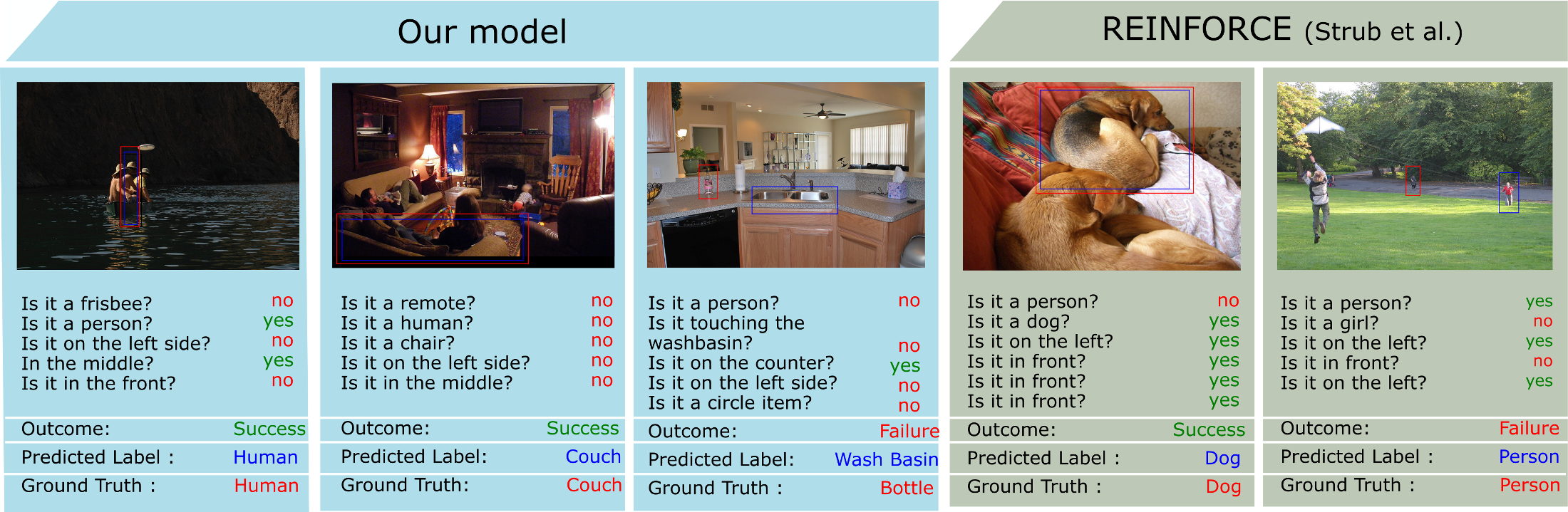}

		\caption{A qualitative comparison of our model with the model proposed by Strub et al. \cite{strub2017end}.
		}
		\label{qualitative}
\end{figure*} 

\subsection {Ablation Study}
We performed an ablation analysis over RIG in order to identify its main learning components. The results of the experiments with the reward function based on RIG are presented in Table~\ref{table3}, whereas Table~\ref{table4} compares the different components of RIG when used as a loss function. The results mentioned under \textit{New Images} refer to images in the test set, while the results shown under \textit{New Objects} refer to the analysis made on the training dataset with different undisclosed objects from the ones used during training time. For the first set of experiments, we compared the performance of information gain vs.\ RIG with the skewness coefficient for goal-oriented visual question generation. It is possible to observe that RIG is able to achieve an absolute improvement of 10.57\% over information gain when used as a reward function and a maximum absolute improvement of 2.8\% when it is optimized in the loss function. 
Adding the skewness term results in a maximum absolute improvement of 0.9\% for the first case and an improvement of 2.3\% for the second case. Furthermore, we compared the performance of the model when trained using RIG but without policy gradients. The model then achieves an improvement of 10.35\% when information gain is used as a loss function. 



\subsection {Qualitative Analysis}
In order to further analyze the performance of our model, we assess it in terms repetitive questions, since they compromise the framework's efficiency. We compare our model with the one proposed by~\cite{strub2017end} and calculate the average number of repetitive questions generated for each dialogue. The model by Strub et al. achieved a score of 0.82, whereas ours scored 0.36 repeated questions per dialogue and 0.27 using RIG as a reward function.

\section{Discussion}

Our model was able to achieve an accuracy of 67.19\% for the task of asking goal-oriented questions on the GuessWhat?! dataset. This result is the highest obtained so far among existing approaches on this problem, albeit still far from human-level performance on the same task, reportedly of 84.4\%. Our gains can be explained in part by how RIG with the skewness component for goal-oriented VQG constrains the process of generating relevant questions and, at the same time, allows the agent to reduce the search space significantly, similarly to decision trees and reinforcement learning, but in a very challenging scenario, since the search space in generative models can be significantly large.

\begin{table}

    \begin{center}
    \scalebox{0.7}{
        \begin{tabular}{|l|c|c|}
        
        \hline
        {\textbf{Rewards}} &{\textbf{ New } } &{\textbf{ New  } } \\
        \raisebox{0ex}{} &\raisebox{0ex}{\textbf{ Images } } &\raisebox{0ex}{\textbf{ Objects } }\\
        \hline
         
        \small I.G. (greedy)& {51.6\%} & {52.4\%}\\\hline
        \small I.G. + skewness (greedy) &{57.5\%} &{62.4\%} \\\hline
        \small R.I.G. (greedy) &{\textbf{58.8\%}} &{\textbf{63.03\%}} \ \\\hline
        
        \hline
        
        \end{tabular}}
    
    \end{center}
    
    \caption{An ablation analysis using Regularized Information Gain as a reward on the GuessWhat?! dataset. }
    \label{table3}
\end{table}

\begin{table}

    \begin{center}
    \scalebox{0.65}{
    
        \begin{tabular}{|l|c|c|}
        \hline
        {\textbf{Approach}} &{\textbf{ New } } &{\textbf{ New  } } \\
        \raisebox{0ex}{} &\raisebox{0ex}{\textbf{ Images } } &\raisebox{0ex}{\textbf{ Objects } }\\
        \hline
         
        I.G. as a loss function & {51.2\%} &{52.8\%} \\
        with no rewards & & \\\hline
        I.G. as a loss function & {57.3\%} & {61.9\%}\\
        with 0-1 rewards (greedy) & & \\\hline
        I.G. + skewness as a loss function  & {59.47\%} & {62.44\%}\\
        with 0-1 rewards (greedy) & &\\\hline
        R.I.G. as a loss function & {\textbf{60.18\%}} &{\textbf{63.15\%}} \\
        with 0-1 rewards (greedy) & & \\\hline

        \end{tabular}}
    \end{center}

    \caption{An ablation analysis of using Regularized Information Gain as a loss function with 0-1 rewards. The figures presented in the table indicate the accuracy of the model on the GuessWhat?! dataset.}
    \label{table4}
\end{table}


Our qualitative results also demonstrate that our approach is able to display certain levels of strategic behavior and mutual consistency between questions in this scenario, as shown in Figure~\ref{qualitative}. The same cannot be said about previous approaches, as the majority of them fail to avoid redundant or other sorts of expendable questions. We argue that our cost function and the skewness coefficient both play an important role here, as the former penalizes synonymic questions and the latter narrows down the set of optimal questions.

Our ablation analysis showed that information gain alone is not the determinant factor that leads to improved learning, as hypothesized by Lee et al.~\cite{lee2018answerer}. However, Regularized Information Gain does have a significant effect, which indicates that a set of constraints, especially regarding the cost of making a question, cannot be taken lightly in the context of goal-oriented VQG. 




\section {Conclusion}
In this paper we propose a model for goal-oriented visual question generation using two different approaches that leverage information gain with reinforcement learning. Our algorithm achieves improved accuracy and qualitative results in comparison to existing state-of-the-art models on the GuessWhat?! dataset. We also discuss the innovative aspects of our model and how performance could be increased. 
Our results indicate that RIG is a more promising approach to build better-performing agents capable of displaying strategy and coherence in an end-to-end architecture for Visual Dialogue.

\section*{Acknowledgments}
We acknowledge partial support of this work by the S\~ao Paulo Research Foundation (FAPESP), grant 2015/26802-1.

\bibliography{acl2019}
\bibliographystyle{acl_natbib}

\end{document}